\documentclass[11pt]{article}

\usepackage{microtype} 
\usepackage{booktabs}  
\usepackage{url}  

\usepackage{amsmath}
\usepackage{amsthm}
\usepackage{graphicx}
\usepackage{amsmath}
\usepackage{amssymb}
\usepackage{booktabs}
\usepackage{xcolor}
\usepackage{comment}
\usepackage{algorithm}
\usepackage{algpseudocode}

\usepackage[final,workshop]{automl}
\usepackage[workshop]{automl}

\usepackage{natbib}
\title{Distilled Pruning: Using Synthetic Data to Win the Lottery}

\author[1]{\nameemail{Luke McDermott}{luke@modernintelligence.ai}}
\author[1]{\nameemail{Daniel Cummings}{daniel@modernintelligence.ai}}

\affil[1]{Modern Intelligence}

\hypersetup{%
  pdfauthor={Luke McDermott, Daniel Cummings}, 
  pdftitle={Distilled Pruning: Using Synthetic Data to Win the Lottery},
  pdfkeywords={Neural Network Pruning, Data Distillation, Neural Architecture Search}
}

\begin{document}

\maketitle

\begin{abstract}
This work introduces a novel approach to pruning deep learning models by using distilled data. Unlike conventional strategies which primarily focus on architectural or algorithmic optimization, our method reconsiders the role of data in these scenarios. Distilled datasets capture essential patterns from larger datasets, and we demonstrate how to leverage this capability to enable a computationally efficient pruning process. Our approach can find sparse, trainable subnetworks (a.k.a. lottery tickets) up to 5x faster than Iterative Magnitude Pruning at comparable sparsity on CIFAR-10. The experimental results highlight the potential of using distilled data for resource-efficient neural network pruning, model compression, and neural architecture search.
\end{abstract}

\section{Introduction}
As prevalent types of deep learning models continue to grow in size and scale, the study of model compression techniques continues to be vitally essential as it addresses the issues of cost-effectiveness, limited computational resources, and model complexity or latency. 
One key capability in this field, neural network pruning \citep{pruning1, pruning2}, has naturally risen in popularity
as it aims to prune or cut out unnecessary parameters in models. Early pruning literature believed that, while dense, overparameterized models are important for training, they are not necessary for inference. This led pruning to be viewed as a post-training procedure, focusing on efficiency of models at inference. \cite{LotteryTicketHypothesis} have shown that this is not the case, emphasizing the potential for pruning at initialization. The Lottery Ticket Hypothesis states that sparse, trainable subnetworks exist at initialization within these dense, overparameterized neural networks. To find these subnetworks or lottery tickets, Iterative Magnitude Pruning (IMP) is augmented with weight rewinding. The IMP process iterates by training a network, pruning the lowest magnitude weights, and rewinding the weights to their initial values or to some point early in training. This repeats until the desired sparsity\footnote{We denote sparsity as percentage of parameters pruned.} is achieved. With weight rewinding, IMP requires the use of post-training information to find optimal masks at initialization. This algorithm enables the study of sparse neural architecture \citep{LTH1, LTH2, LTH3, LinearModeConnectivity}, providing a way for researchers to consistently find ``lucky" lottery tickets.

Even as a fundamental research tool, IMP is largely inefficient due to the extensive retraining process. To achieve some sparse model with IMP, one must retrain some network numerous times over to achieve the mask, then retrain one final time to validate the sparsity mask. To address this issue, we employ the same framework as IMP, but instead use distilled data \citep{DataDistillation}, essentially a summarized version of our training data, in the inner training loop to approximate trained weights. As a result, sparsity masks can be generated in considerably less time while still being capable of achieving full accuracy when trained with real training data. We show in our setting that distilled data can pick winning tickets.

Previous work, such as \cite{DataDiet}, demonstrated that subsets of the training data are sufficient for finding lottery tickets. We improve upon this idea by distilling the essential features of a class into a few synthetic images. Data distillation condenses a dataset into a small, synthetic sample, which, when used for training, yields similar performance to training on the real dataset. Often, this means reducing a dataset to 1, 10, or 50 images per class. This topic has seen rapidly growing interest due to the benefits of lower computational overhead for model training and can broadly be separately into the subcategories of meta-model matching, gradient matching, distribution matches, and trajectory matching \citep{sachdeva2023data, FRePo, MTT, RCIG, KIP}.

A downside of state-of-the-art data distillation methods is that they require significant memory overhead which limits their ability to scale to larger model sizes (and thus lack of cross-architecture generalizability). Recent works like \cite {RCIG} and \cite{FRePo} have explored the transferrability of datasets generated by such methods on ResNet \citep{resnet} and VGG \citep{vgg}, but since these results leave a lot of room for improvement, we focus our work to more computationally tractable convolutional networks. Despite concerns of distillation methods, in our unique setting with heavy retraining in IMP, poorly-generalizing distillation methods still show substantial utility in improving the retraining process since we only have to optimize the distilled data for one model family.

In this paper, we introduce data distillation as a means to accelerate retraining in iterative pruning methods, while still accurately identifying winning tickets for the original dataset. We emphasize the use for distilled pruning as a means of rapid experimentation in pruning and NAS research, taking advantage of the efficiency/performance trade off. Data distillation and neural network nruning can be viewed as orthogonal approaches to computational efficiency, so data distillation provides additional speed up in retraining that can be used with other efficient pruning methods, not just IMP.
\section{Method}
Formally, the Lottery Ticket Hypothesis \citep{LotteryTicketHypothesis} conjectures that for some randomly initialized, dense neural network $f(x;\theta)$, there exists a non-trivial binary mask $m \in \{0,1\}^{|\theta|}$, such that when trained in isolation on some training data $D_\text{train}$, the subnetwork $f(x; \text{train}(\theta \odot m, D_\text{train}))$ achieves similar performance to $f(x; \text{train}(\theta, D_\text{train}))$. We denote $\odot$ as elementwise multiplication and assume there exists some sufficient SGD-based train function, $\textit{train}:\; \mathbb{R}^{|\theta|} \rightarrow \mathbb{R}^{|\theta|}$. To find such $m$, pruning researchers employ IMP as follows: 1) Train the network for $n$-epochs, 2) remove 20\% of the non-pruned weights prioritizing by lowest magnitude, 3) rewind the weights back to initialization or some early point in training, 4) Iterate Steps 1-3 until desired sparsity. Here, sparsity is defined as the percentage of parameters pruned. 

We employ a simple augmentation to the original IMP algorithm by replacing the training data, $D_{\text{train}}$, needed to find the sparsity mask with distilled data, $D_{\text{syn}}$, as demonstrated by the Algorithm \ref{alg:alg1}. We also train for some $t$-many epochs on the distilled data, while preserving the $n$-long training with real data at the end. The source of distilled data is largely plug-and-play, and we encourage researchers and practitioners alike to use the most applicable distillation method that fits their performance needs and computational budget. In future work, we plan to benchmark across different data distillation methods.

\begin{algorithm}
\caption{Distilled Pruning}\label{alg:alg1}
\begin{algorithmic}
\State \textbf{Inputs}: $\theta_{\text{init}}, D_\text{syn}, D_\text{real}, \text{desired sparsity, amount}$
\State $\theta \gets \theta_{\text{init}}$
\State $m \gets \mathbf{1}$ \Comment{Initialize as matrix of 1's of size $|\theta|$ }
\While{$sparsity(m) < \text{desired sparsity}$}
    \State{$\theta \gets train(\theta \odot m, D_\text{syn}, t \text{ epochs})$}
    \State{$m \gets prune(\theta \odot m \text{, amount})$}
    \State{$\theta \gets \theta_{\text{init}}$}
\EndWhile
\State{$\theta_{\text{finetune}} \gets train(\theta \odot m, D_\text{real}, n \text{ epochs})$}\\
\Return{$\theta_{\text{finetune}}, m$}
\end{algorithmic}
\end{algorithm}

Specifically for our experiments, we utilize MTT as demonstrated by \cite{MTT} due to ease of reproducibility. MTT leverages the concept of expert trajectories, which are snapshots of parameters from models trained on the real dataset. The goal is to induce a similar trajectory in the student model trained on synthetic data, leading to similar test performance. We refer the reader to the original paper for implementation level details. 
\section{Experiments}
For our experiments, we chose AlexNet \citep{AlexNet} for CIFAR-10 \citep{CIFAR} and a 128-width ConvNet for CIFAR-100 \citep{CIFAR} to maintain consistency with experiments in previous literature by \cite{MTT}. We distilled each class down to 10 or 50 images, denote as 10 ipc (images per class)  or 50 ipc. The distilled CIFAR-10 has a size of 100 or 500 training images and 1,000 or 5,000 for CIFAR-100.
\subsection{Sparsity Analysis}
\begin{figure}
    \centering
    \makebox[\textwidth][c]{\includegraphics[width=1.1\linewidth]{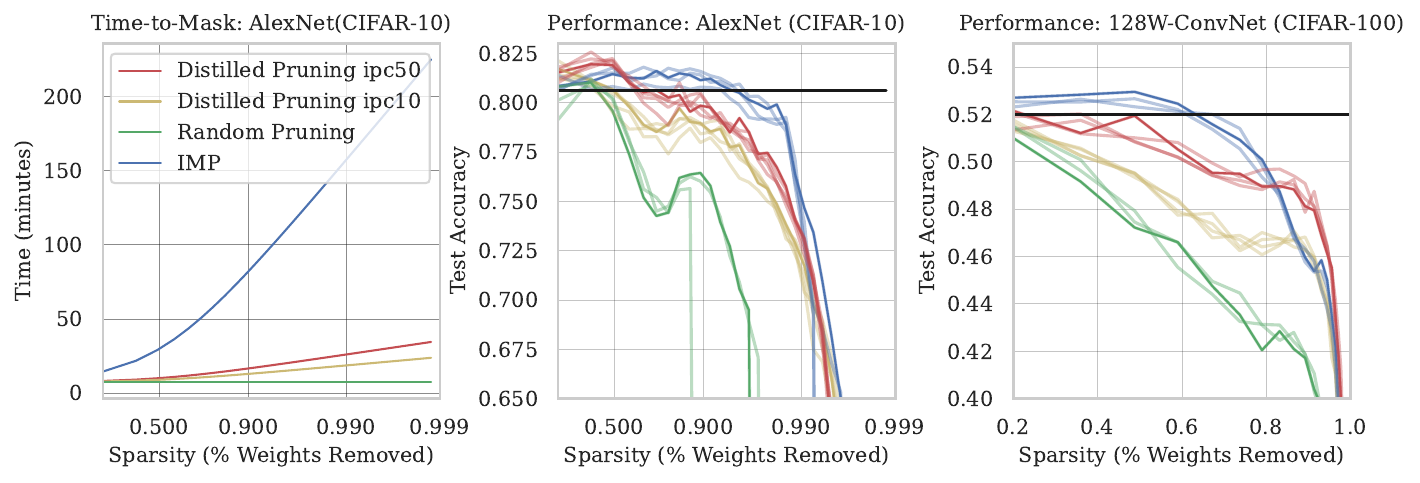}}%
    \caption{Sparsity Mask performance for AlexNet on CIFAR-10 and a 128-width ConvNet on CIFAR-100 across methods. Best seed of each method is bolded. We pruned 20\% of weights at each iteration up to 30 iterations for CIFAR-10 and 20 for CIFAR-100. Random mask selects weights at random each iteration. lottery tickets exist if test accuracy of sparse model achieves or surpasses the dense model accuracy as shown in black. Time-to-mask measured by time to prune and retrain the sparsity mask with real data.}
    \label{fig:sva}
\end{figure}
In our setting, distilled data is appropriately finding lottery tickets at non-trivial sparsities, showing that at 50 ipc the approximated weights from distilled training are sufficient for IMP. Figure \ref{fig:sva} shows we achieve relatively comparable performance to IMP at mid to high sparsities and even outperform at low sparsities for CIFAR-10. For CIFAR-100, we see a fall off earlier as Distilled Pruning finds lottery tickets up to only 50\% sparsity. For both datasets, 10 ipc performs poorly as expected due to low performance even on data distillation objectives \citep{MTT, FRePo, RCIG, KIP}. We believe with the current state of data distillaton methods, Distilled Pruning may not scale to deeper networks or to datasets with high amounts of outliers yet. As a rapidly evolving field, we expect this to change soon as the field matures. 
\subsection{Efficiency Analysis}

In Figure \ref{fig:sva}, we present compelling evidence showcasing the significant speedup achieved with distilled pruning compared to standard IMP. Measured on an Nvidia RTX A4000 GPU, we achieve an average of 55 seconds per distilled training session on CIFAR-10 distilled to 50 ipc, compared to 7.25 minutes per training on real data. Distilled Pruning found a lottery ticket of comparable accuracy at roughly 90\% sparsity in CIFAR-10, resulting in a 5x speed up. While distilled pruning with ipc10 looks useful here, the performance drop off is too large for the minimal improvement in time-to-mask. It is worth noting that the major computational burden associated with distilled pruning lies in the final retraining phase using real data. Consequently, in scenarios where validation of a sparsity mask is unnecessary, distilled pruning enables us to generate masks 8 times faster than with IMP. 

One of the key advantages of distilled pruning is the ability to rapidly prototype and experiment, particularly for researchers working within a limited set of datasets or compute resources. Synthetic data is generated once per dataset, providing a means for quick and convenient experimentation. Moreover, synthetic data is pre-computed and publicly available for popular datasets, further streamlining the research process. By capitalizing on the plug-and-play nature of distillation methods, any advancements in data distillation techniques can directly translate into speed improvements for distilled pruning. Because of this, we exclude time-to-distill from our plot. For reference, MTT has one of the largest computational costs for distillation, but only takes an additional 133 minutes to distill CIFAR-10 to 50 images per class \citep{MTT}. We emphasize that for popular datasets, these are often pre-computed and publicly available with state-of-the-art distillation methods.

\subsection{Instability Analysis}
\begin{figure*}
    \centering
    \includegraphics[width=1\textwidth]{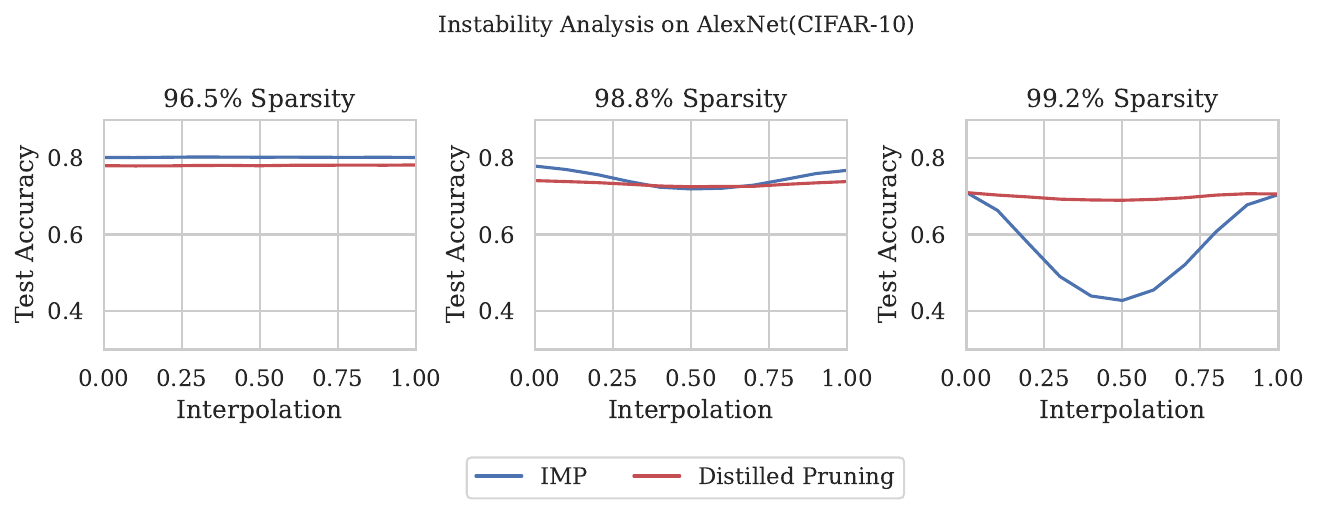}
    \caption{The test accuracy for interpolated weights between two models trained with different SGD noise with AlexNet and CIFAR-10. Each plot uses a fixed sparsity mask found by IMP or Distilled Pruning. A drop in accuracy implies no linear mode connectivity or instability to SGD noise. Distilled Pruning uses 50 images per class.}
    \label{fig:lmc}
\end{figure*}
To gain deeper insights into the distinctions between winning tickets obtained through distilled pruning and those discovered using IMP, we employ an \textit{instability analysis} inspired by \cite{LinearModeConnectivity}. 
 As described in previous literature, lottery tickets exhibit linear mode connectivity, representing stability to noise from stochastic gradient descent (SGD). In Figure \ref{fig:lmc}, the selected sparsity masks were trained using two distinct permutations of real training data. Then, we performed a linear interpolation between the trained weights of the two networks. This process allowed us to observe the linear mode connectivity and assess the stability of the models. A drop in test accuracy during this interpolation means the model is unstable.

We observed distinctions in the lottery tickets yielded through the two methods. In the case of IMP-generated subnetworks, we observed the need for rewinding to an early point in training (specifically, after one epoch, $k=1$) as opposed to initialization,  aligning with previous work \citep{LinearModeConnectivity}. In contrast, the lottery tickets identified through distilled pruning proved to be drastically more stable against SGD noise, not requiring any rewinding. We found our tickets maintained linear mode connectivity at extreme sparsities, only falling during model collapse.

These discoveries hint at the possibility of distilled pruning producing a different type of lottery ticket, where trained weights approximated by distilled data might provide unique and valuable insights into the Lottery Ticket Hypothesis. However, as \cite{LinearInterpolation} discuss, data augmentation, initializations, and optimizers all play significant roles in linear interpolation. Even then, they show stability does not always predict test accuracy. Therefore, we believe further research is necessary to fully understand why distilled data-generated tickets exhibit such stability, even when rewound to initialization.

\section{Conclusion}
In this pilot study, we explore the effect of data distillation on neural network pruning. The implications of distilled pruning extend far beyond its direct applications. Fast prototyping becomes more accessible for researchers who leverage pruning techniques, as the distilled pruning framework enables swift iteration and experimentation with various pruning configurations. Additionally, distilled pruning serves as a valuable tool for Neural Architecture Search (NAS) validation, facilitating the assessment of architectures' performance and characteristics. One notable advantage within pruning is the flexibility it provides in terms of pruning granularity. Researchers can substantially increase in the number of pruning iterations, allowing for pruning of smaller amounts of weights per iteration. This hyper-iterative approach grants precise control over the levels of pruning, enabling fine-grained exploration of the sparsity spectrum. Distilled pruning effectively reduces the sample complexity of mask generation, thereby opening up new avenues for stochastic approaches to IMP or even larger-scale NAS methods.

While our research focuses on highlighting the speed-up achieved with distilled pruning, we acknowledge that there is a trade-off in performance compared to the standard IMP method. As data distillation as a field matures, we expect to close the performance gap and apply this method to larger models and datasets. In future work, we plan to test a wider range of novel distillation methods such as \cite{FRePo} and \cite{RCIG} while exploring the scalability of distilled pruning with larger architectures. 

\section{Broader Impact Statement}

Our proposed solution employs distilled data, which leads to significant computational savings during the pruning process. This reduction in computational requirements directly translates to diminished CO2 emissions, contributing to more sustainable AI research and development practices. Our approach would generalize well to models outside computer vision and could lead to more effective and efficient pruning solutions in areas such as natural language processing, and generative architectures. Moreover, this work can make advanced neural network design more accessible to a broader range of researchers and developers, reducing the expertise and compute infrastructure required to prune high-performing networks. 

One possible risk of our approach is the loss of detail when using distilled data. While data distillation aims to retain as much useful information as possible, there's a risk that some important outlier data could be lost in the process, potentially leading to unexpected model performance or biased outcomes. To counter the potential risks associated with data distillation, it's crucial to validate distilled datasets thoroughly against real-world data to ensure they adequately represent the problem space. While data distillation holds considerable promise for the future enhancement of deep learning, the drawbacks and related mitigation strategies should continue to be carefully studied.  

%
%
%
%
%
\newpage
{\small
\bibliography{egbib}

\begin{thebibliography}{}

\bibitem[Cazenavette et~al., 2022]{MTT}
Cazenavette, G., Wang, T., Torralba, A., Efros, A.~A., and Zhu, J.-Y. (2022).
\newblock Dataset distillation by matching training trajectories.
\newblock {\em CVPR}.

\bibitem[Chen et~al., 2021]{LTH2}
Chen, X., Cheng, Y., Wang, S., Gan, Z., Liu, J., and Wang, Z. (2021).
\newblock The elastic lottery ticket hypothesis.
\newblock {\em NIPS}.

\bibitem[Frankle and Carbin, 2019]{LotteryTicketHypothesis}
Frankle, J. and Carbin, M. (2019).
\newblock The lottery ticket hypothesis: Finding sparse, trainable neural
  networks.
\newblock {\em ICLR}.

\bibitem[Frankle et~al., 2020]{LinearModeConnectivity}
Frankle, J., Dziugaite, G.~K., Roy, D.~M., and Carbin, M. (2020).
\newblock Linear mode connectivity and the lottery ticket hypothesis.
\newblock {\em PMLR}.

\bibitem[Han et~al., 2015]{pruning2}
Han, S., Pool, J., Tran, J., and Dally, W.~J. (2015).
\newblock Learning both weights and connections for efficient neural networks.
\newblock {\em NIPS}.

\bibitem[He et~al., 2015]{resnet}
He, K., Zhang, X., Ren, S., and Sun, J. (2015).
\newblock Deep residual learning for image recognition.

\bibitem[Krizhevsky, 2009]{CIFAR}
Krizhevsky, A. (2009).
\newblock Learning multiple layers of features from tiny images.

\bibitem[Krizhevsky et~al., 2017]{AlexNet}
Krizhevsky, A., Sutskever, I., and Hinton, G.~E. (2017).
\newblock Imagenet classification with deep convolutional neural networks.
\newblock {\em ACM}, 60(6):84–90.

\bibitem[Lecun et~al., 1989]{pruning1}
Lecun, Y., Denker, J., and Solla, S. (1989).
\newblock Optimal brain damage.
\newblock {\em NIPS}, 2:598--605.

\bibitem[Loo et~al., 2023]{RCIG}
Loo, N., Hasani, R., Lechner, M., and Rus, D. (2023).
\newblock Dataset distillation with convexified implicit gradients.

\bibitem[Ma et~al., 2021]{LTH3}
Ma, X., Yuan, G., Shen, X., Chen, T., Chen, X., Chen, X., Liu, N., Qin, M.,
  Liu, S., Wang, Z., and Wang, Y. (2021).
\newblock Sanity checks for lottery tickets: Does your winning ticket really
  win the jackpot?
\newblock {\em NIPS}.

\bibitem[Nguyen et~al., 2021]{KIP}
Nguyen, T., Chen, Z., and Lee, J. (2021).
\newblock Dataset meta-learning from kernel ridge-regression.

\bibitem[Paganini and Forde, 2020]{Pretoporter}
Paganini, M. and Forde, J.~Z. (2020).
\newblock Bespoke vs. pr\^et-\`a-porter lottery tickets: Exploiting mask
  similarity for trainable sub-network finding.

\bibitem[Paul et~al., 2022a]{LTH1}
Paul, M., Chen, F., Larsen, B.~W., Frankle, J., Ganguli, S., and Dziugaite,
  G.~K. (2022a).
\newblock Unmasking the lottery ticket hypothesis: What's encoded in a winning
  ticket's mask?

\bibitem[Paul et~al., 2022b]{DataDiet}
Paul, M., Larsen, B.~W., Ganguli, S., Frankle, J., and Dziugaite, G.~K.
  (2022b).
\newblock Lottery tickets on a data diet: Finding initializations with sparse
  trainable networks.
\newblock {\em NIPS}.

\bibitem[Sachdeva and McAuley, 2023]{sachdeva2023data}
Sachdeva, N. and McAuley, J. (2023).
\newblock Data distillation: A survey.
\newblock {\em TMLR}.

\bibitem[Simonyan and Zisserman, 2015]{vgg}
Simonyan, K. and Zisserman, A. (2015).
\newblock Very deep convolutional networks for large-scale image recognition.

\bibitem[Vlaar and Frankle, 2022]{LinearInterpolation}
Vlaar, T. and Frankle, J. (2022).
\newblock What can linear interpolation of neural network loss landscapes tell
  us?
\newblock {\em ICML}.

\bibitem[Wang et~al., 2020]{DataDistillation}
Wang, T., Zhu, J.-Y., Torralba, A., and Efros, A.~A. (2020).
\newblock Dataset distillation.

\bibitem[Zhou et~al., 2022]{FRePo}
Zhou, Y., Nezhadarya, E., and Ba, J. (2022).
\newblock Dataset distillation using neural feature regression.
\newblock {\em NIPS}.

\end{thebibliography}
}
\newpage
\appendix
\section{Reproducibility}
We adapted our code from MTT and built a pruning framework on top of it. For review, our code is available: https://github.com/luke-mcdermott-mi/distilled-pruning. MTT can be found at  https://github.com/GeorgeCazenavette/mtt-distillation. We used the pre-computed distilled data from the original paper. For training on distilled data we used the default hyperparameters in the MTT respository, but used a learning rate of .01 and .007 for the 50 ipc (images per class) and 10 ipc CIFAR-10 datasets respectively. We also used 1000 and 3000 epochs for these. For CIFAR-100, we used a learning rate of .01 and .09 for the 50 ipc and 10 ipc datasets respectively with 1250 epochs of training. For training on real data, we used a .0008 learning rate, 512 batch size, .0008 weight decay, and gamma of .15. Milestones were placed at epochs 50,65, and 80. We found AlexNet only needed 60 epochs on CIFAR-10 which provides a stronger baseline in terms of time-to-mask. Any additional training time found negligable performance. Furthermore, we used 120 epochs for the 128-Width ConvNet on CIFAR-100. We tuned hyperparameters using Optuna and spent roughly a half gpu-hour for distilled tuning and 2 gpu-hours for tuning on the original datasets on an Nvidia A6000. These hyperparameters can also be found in our codebase. For random seeds, we used seeds 0-4. 
\section{Weight Distribution of Sparsity Masks}
\begin{figure*}
    \makebox[\textwidth][c]{\includegraphics[width=1.25\textwidth]{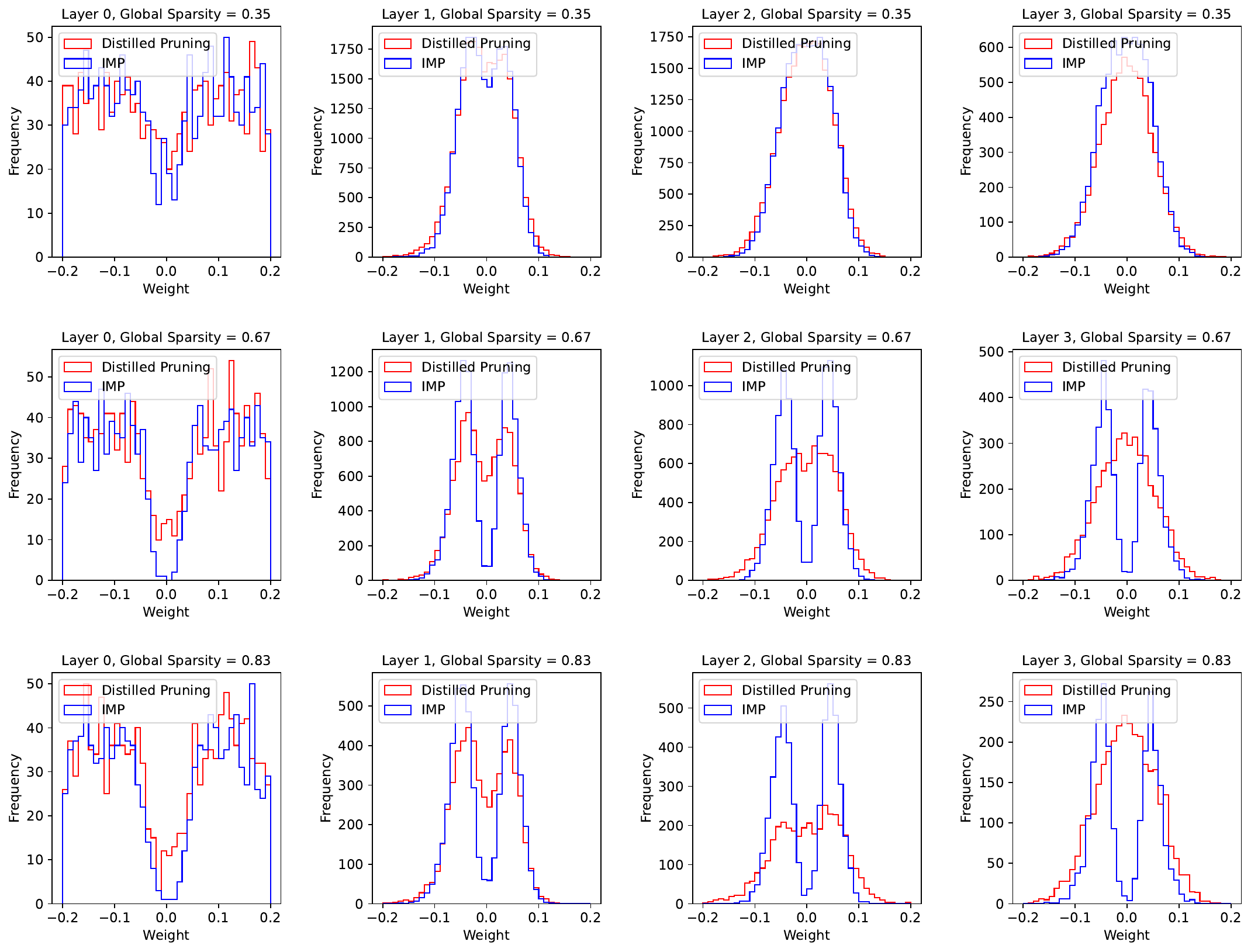}}%
    \caption{Initialization values of weights for lottery tickets found on a ConvNet \& CIFAR-10. Going from left to right, we look at deeper layers of the ConvNet: three convolutional feature extractors and a final linear classifier. Going from top to bottom, we view sparser lottery tickets.}
    \label{fig:weights}
\end{figure*}
Following \cite{Pretoporter}, we employ a similar analysis of our lottery tickets found across both pruning methods in figure \ref{fig:weights}. We showcase the distribution of the non-pruned initialized weights. Traditional lottery tickets exhibit a bimodal distribution at high sparsities which can be confirmed by our findings. Essentially, weights that are initialized near zero are likely to stay near zero after training with IMP. We find that in a small ConvNet (3 convolutional layers + 1 linear layer) on CIFAR-10, distilled pruning agrees with IMP on the distribution of layer sparsity; however, they disagree on weight distribution at later layers, especially at high sparsities. These insights point to more distinctions between lottery tickets that we could not cover in our instability analysis. For this experiment, we used IMP with rewinding back to initialization. We found in the case with AlexNet, which required rewinding to the first epoch in training, that even poorly trained weights show a drastically different distribution than initialization. This made it difficult to compare Distilled Pruning and IMPs' weights on AlexNet as their rewind weights were different.
\begin{acknowledgements}

\end{acknowledgements}





\end{document}